\documentclass[letterpaper, 10 pt, conference]{ieeeconf}  

\IEEEoverridecommandlockouts       

\usepackage{amsmath,amsfonts}
\usepackage{algorithmic}
\usepackage{algorithm}
\usepackage{array}
\usepackage{textcomp}
\usepackage{stfloats}
\usepackage{url}
\usepackage{verbatim}
\usepackage{graphicx}
\usepackage{balance}
\usepackage[font=small,labelfont=bf,justification=centering,labelsep=period]{caption}
\usepackage[font=normalsize,labelfont=sf,textfont=sf]{subcaption}
\usepackage{siunitx}
\usepackage{xcolor}
\usepackage{soul}
\usepackage{lipsum} 
\usepackage{changepage}
\usepackage{makecell}
\usepackage{booktabs}
\usepackage{adjustbox}
\usepackage{multicol}
\usepackage{color}
\usepackage{diagbox}
\usepackage{siunitx}
\usepackage{multirow}

\graphicspath{{./Figures/}}
\DeclareGraphicsExtensions{.pdf,.jpeg,.png,.pdf_tex}

\title{\LARGE \bf
Friction Characterization of a Cable-Driven Differential \\ Actuation System for Lower-Limb Exoskeletons*}

\author{Alberto Maria Nobili$^{1}$, Fabio Salsedo$^{2}$, Alessandro Filippeschi$^{1}$
\thanks{*This project has received funding from the European Union’s Horizon Europe programme under the Marie Skłodowska-Curie Grant Agreement No. 101169197 through the AERIALIST Doctoral Network.}
\thanks{$^{1}$Alberto Maria Nobili and Alessandro Filippeschi are with the Institute of Mechanical Intelligence and with the Department of Excellence in Robotics and AI, Sant'Anna School of Advanced Studies, Pisa, Italy
        {\tt\small albertomaria.nobili@santannapisa.it, alessandro.filippeschi@santannapisa.it}}%
\thanks{$^{2}$Fabio Salsedo is with Wearable Robotics s.r.l., Pisa, Italy
        {\tt\small f.salsedo@wearable-robotics.com}}%
}

\begin{document}

\maketitle
\thispagestyle{empty}
\pagestyle{empty}

\begin{abstract}

Lower-limb exoskeletons require actuation systems that can provide accurate joint torque control while preserving low mass and encumbrance. Conventional architectures often rely on independently actuated joints and joint-level torque sensors, increasing system complexity and weight. This paper presents a novel differential actuation architecture for hip–knee flexion/extension, enabling cooperative torque sharing between two motors via a linear differential mapping between motor and joint. To compensate for transmission losses, a model-based friction estimation strategy is developed and experimentally implemented, allowing accurate joint torque estimation without the need for torque sensors. The proposed solution is validated on a physical prototype, demonstrating the feasibility of sensorless torque estimation in a differentially actuated hip–knee module of a lower-limb exoskeleton.

\end{abstract}
\section{Introduction}
\label{sec:I}

Lower-limb exoskeletons are wearable robotic devices designed to restore, augment, or assist human locomotion in rehabilitation \cite{zhou2021lower}, industrial \cite{kuber2023systematic}, and military \cite{jia2020preliminary} applications. The performance of such systems depends on both the actuation architecture and the ability to provide accurate torque control at each joint.

Traditional exoskeletons most commonly employ independent actuators per joint, typically based on electric motors combined with rigid transmissions such as harmonic gearboxes \cite{bettella2025scoping, choi2019control} or compliant elements such as series elastic actuators (SEAs) \cite{yu2023design, liu2024synergetic}. However, they are usually joint-collocated and introduce additional mass and volume, which can be particularly limiting in multi-joint portable exoskeletons.
Alternative transmission solutions, such as quasi-direct drives (QDDs) \cite{yu2020quasi} and Bowden-cable-based actuation \cite{witte2020design, sun2025design}, have been proposed to reduce reflected inertia or distal mass. 
While these approaches can improve transparency, they present additional challenges. In particular, QDDs do not meet the requirements of lower-limb exoskeleton joints in terms of mass, torque, and speed \cite{xu2025survey}, while Bowden-cables are often affected by significant nonlinear friction, which is a major source of hysteresis and phase delay of the transmission system \cite{shi2022learning}. 

Differential actuation schemes have been extensively studied in robotics to enable cooperative torque generation, with notable applications in robotic wrists \cite{chishty2021kinematic} and elbow \cite{chen2019elbow} exoskeletons. In these systems, differential mechanisms allow multiple actuators to contribute to multiple degrees of freedom, improving efficiency and adaptability to task-dependent load distributions. In contrast, their application to lower-limb exoskeletons remains relatively unexplored, especially for architectures that simultaneously actuate the hip and knee flexion-extensions.

Beyond mechanical design, from a control perspective, several works have addressed joint torque estimation without direct torque sensors, relying instead on motor current measurements combined with dynamic and friction models \cite{devzman2022mechanical, andrade2021role, hassen2021exoskeleton}. Accurate friction modeling has been shown to be critical for achieving reliable torque control, particularly in systems with complex transmissions involving gears, pulleys, or cable routings \cite{baud2021review}. Nevertheless, most existing approaches focus on independently actuated joints and do not explicitly consider the coupling effects introduced by differential transmissions.

The main contributions of this work are: 1) a differential actuation architecture for hip–knee flexion/extension; and 2) a friction modeling and estimation framework specifically developed for the proposed coupled transmission, enabling accurate joint torque estimation without joint-level torque sensors. The proposed approach evaluates the friction losses of the individual transmission branches, accounting for both load-independent and load-dependent effects. For this latter contribution, we assume that the transmission system is composed of gears and tendon-driven elements. This approach contributes to bridging the gap between mechanical torque-sharing concepts and sensor-sparing control strategies, contributing a novel solution tailored to the specific demands of lower-limb exoskeletons

The outline of this article is organized as follows. Section~\ref{sec:II} describes the proposed device's mechanical design. Section~\ref{sec:III} contains the transmission friction model used in Section~\ref{sec:IV} to present the identification procedure. The results of this characterization are presented in Section~\ref{sec:V} and discussed in Section~\ref{sec:VI}. Ultimately, Section~\ref{sec:VII} contains the drawn conclusions about the study and the ideas for future works.

\section{System Description}
\label{sec:II}

\subsection{Exoskeleton General Design}
\label{sec:general_design}

\begin{figure*}
    \centering
    \begin{subfigure}{0.97\columnwidth}
        \centering
        \includegraphics[width=\textwidth]{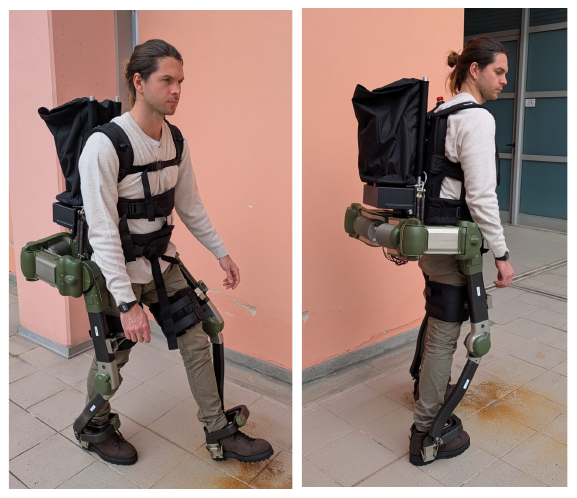}
        \caption{}
        \label{fig:exos_views}
    \end{subfigure}
    \hspace{0.3cm}
    \begin{subfigure}{0.88\columnwidth}
         \centering
         \includegraphics[width=\textwidth]{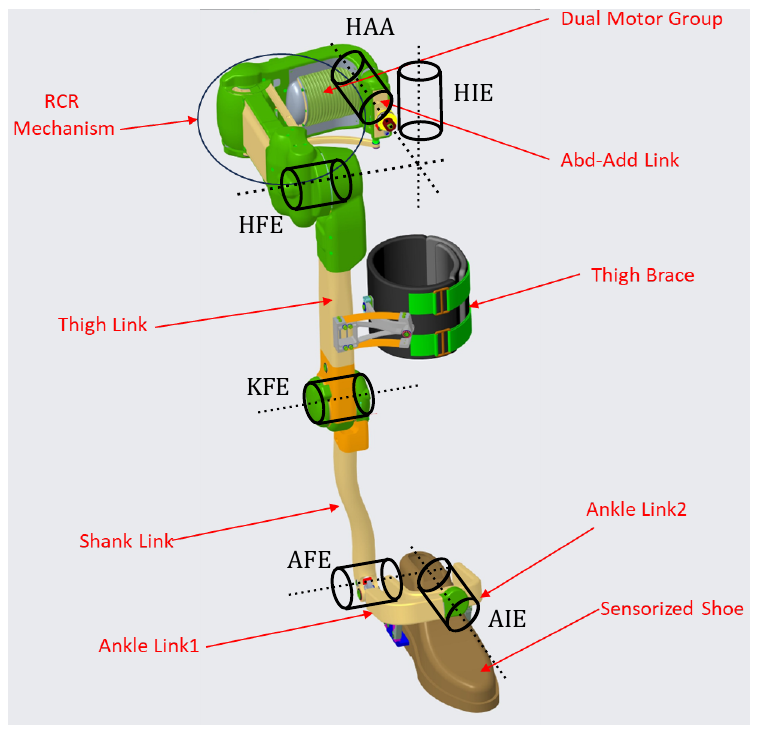}
         \caption{}
         \label{fig:leg_kinematics}
    \end{subfigure}
    \caption{a) Front and back views of the exoskeleton which uses the actuation system described in this study. b) Schematic description of the kinematic chain of an exoskeletal leg, showing the main components.}
    \label{fig:exos_presentation}
\end{figure*}

The proposed actuation system is part of a lower-limb occupational exoskeleton (total weight is \SI{32}{kg}) with two independent 6-DoFs legs (Fig.~\ref{fig:exos_presentation}), developed by Wearable Robotics s.r.l., Pisa, Italy. It is intended to physically support a human subject in carrying equipment/loads located in the backpack, for logistics or similar applications. 
Its design is completely portable, since it features a battery, various local computing units, and a WiFi module for communication with a GUI on a smartphone. Three groups of harnesses are present: one group for the trunk, one for the thighs, and one at the level of the feet. Both the device activation/deactivation and the level of assistance are set through an app on a smartphone.

The exoskeleton presents, on the back, a box containing all the electronic components: a main computing unit (conga-MA3/i-E3815-2G) which controls the legs and monitors the device state, two custom-made motor driver units and two custom-made analog signal acquisition units (all developed with Texas Instruments TMS320F28335 microcontrollers), a \SI{1200}{Wh}-\SI{67}{V} battery, which allows for an autonomy of around 12 hours with a load of \SI{25}{kg} during level walking at \SI{2}{km/h}, and a WI-FI module (VONETS VM300-H) to communicate with the external user interface.

\subsection{Kinematic Structure}
\label{sec:kin_design}


The kinematic structure of each leg is represented by a serial chain of 6 revolute joints (see Fig.~\ref{fig:leg_kinematics}): three for the hip (adduction/abduction HAA, internal/external rotation HIE, and flexion/extension HFE), one for the knee (flexion/extension KFE), and two for the ankle (flexion/extension AFE and inversion/eversion AIE). 

Of the six joints that compose a leg, only 2 are actively actuated (HFE and KFE), while one (HAA) is passively actuated (using mechanical springs), and the others (HIE, AFE, and AIE) are non-actuated. 
This paradigm is particularly suitable for load-carrying tasks, in which the load primarily weighs on the hip and knee flexion-extension plane.


The HIE joint was designed using a patented pantographic structure, which has already been implemented to mimic the shoulder joint in ALEx RS upper limb exoskeleton \cite{ALEx}. With this solution, the combined rotation of the three revolute joints of the HIE implements a remote centre of rotation (RCR).

\subsection{Actuation System}
\label{sec:actuators}

To actuate the two flexion/extensions, two electrical 3-phase brushless motors (Wittenstein MRSx085x-040C \SI{2064}{W}@\SI{60}{V}) are added to each leg, coaxially arranged exploiting a concentric shaft mechanism.
These motors are equipped with encoders for the reconstruction of the angular position $\theta_i$ and, subsequently, of the speed $\dot \theta_i$. Moreover, the phase currents of each motor are measured. In this way, we can extract the quadrature current $Iq_i$ and, therefore, the motor torque $\tau_i=k_m\ Iq_i$ where $k_m$ is the motor constant.

To keep the mass and inertia of the fast-moving links of the exoskeletal legs low, the dual motor group has been located on the first link of the RCR mechanism. 
In this way, the motors are very close to the trunk link of the device, which would be their ideal location.
However, this choice implies the implementation of two transmissions, which can transmit the motor torques to the joints to be actuated (HFE and KFE).

\begin{figure*}
    \centering
    \begin{subfigure}{0.95\columnwidth}
        \centering
        \includegraphics[width=\textwidth]{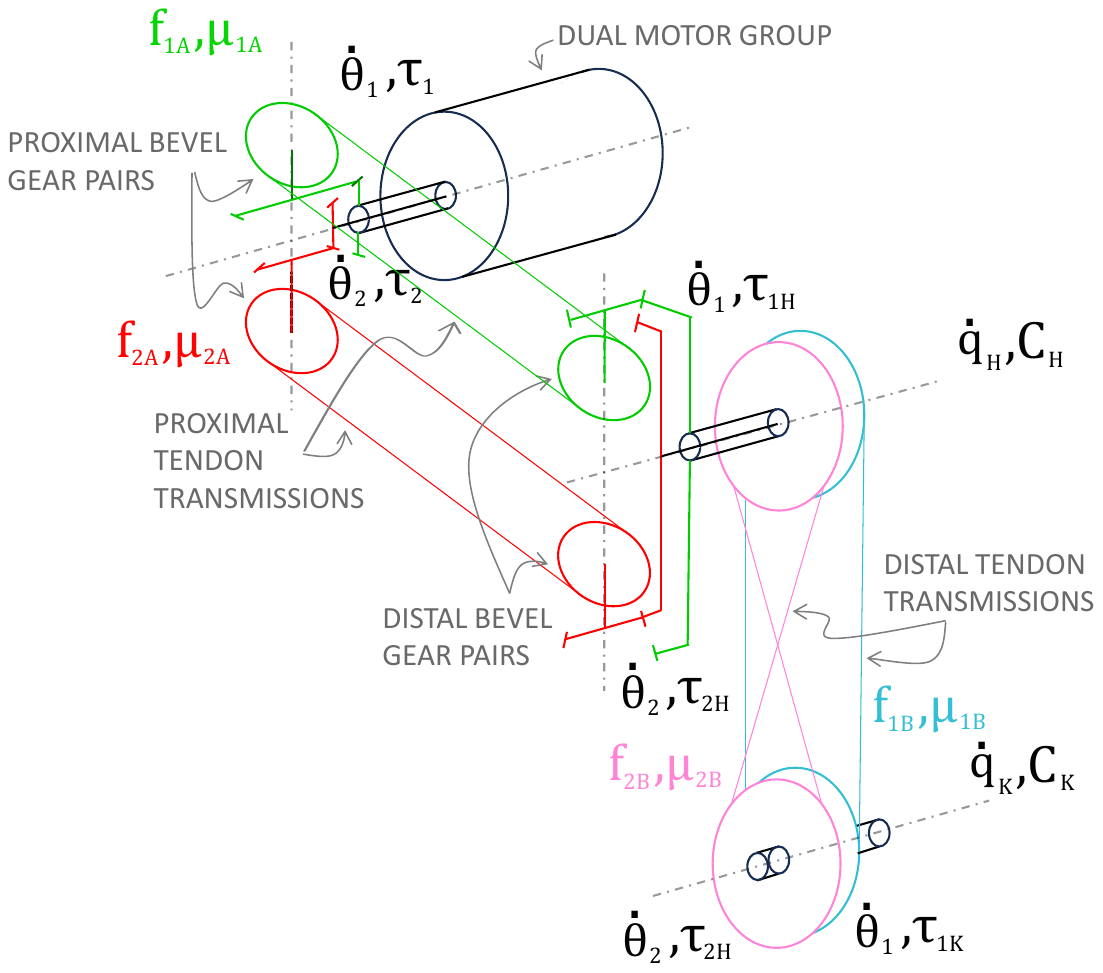}
        \caption{}
        \label{fig:transmission_scheme}
    \end{subfigure}
    \hspace{0.3cm}
    \begin{subfigure}{0.8\columnwidth}
         \centering
         \includegraphics[width=\textwidth]{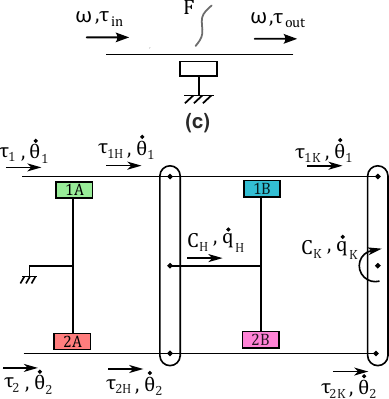}
         \caption{}
         \label{fig:transmission_scheme_model}
    \end{subfigure}
    \caption{a) Schematics of the transmission system used to actuate each leg joint. b) Model used to represent the mechanical transmission. Here, the arrows represent the positive direction of each variable.  c) Scheme used as a model for the friction components acting on each transmission segment.}
\end{figure*}

The transmission system of the exoskeleton is composed of two independent transmission lines, indicated by numbers 1 and 2 in Fig.~\ref{fig:transmission_scheme}. Each line is composed of two stages, indicated by letters A (proximal stage) and B (distal stage) in the same figure. The proximal stage features two pairs of bevel gears, each with its own reduction ratio, and a tendon-driven pulley pair, with no associated speed reduction. The distal stage includes a pair of tendon-driven pulleys, with no associated speed reduction.

In this way, the speed reduction is performed in two different locations and not at one single point, allowing for keeping the loads acting on most of the components of the proximal stage components low and, hence, their encumbrance.
It is also worth noting that the distal transmission stages of the two lines end on driven pulleys, which are rigidly coupled with the shank link, but have different routing. In this way, the torque generated at the hip flexion/extension is the sum of the torques produced by the two motors, while the torque generated at the knee flexion is their difference.

More in detail, the ideal transmission matrix between motor and joint variables is
\begin{equation}\label{eq:transm_matrix_vel}
    \dot q = \frac{1}{2t_r} 
    \begin{bmatrix}
    +1&  +1\\+1& -1    
    \end{bmatrix}
    \dot \theta = \frac{1}{2t_r}\ M \dot{\theta}
\end{equation}
\begin{equation}\label{eq:transm_matrix_tau}
    C = t_r 
    \begin{bmatrix}
    +1&  +1\\ +1& -1    
    \end{bmatrix}
    \tau = t_r\ M\  \tau
\end{equation}
where $t_r=15.125$ is the transmission ratio between motors and joints, $\dot q =  [\dot q_H, \dot q_K]^T$ and $\dot \theta = [\dot \theta_1, \dot \theta_2]^T$ are the speeds of joints and motors, respectively,  $C =  [C_H, C_K]^T $ and $\tau = [\tau_1, \tau_2]^T$ are the torques exerted at the joints and motors, respectively.

\begin{figure}
    \centering
    \includegraphics[width=1\columnwidth]{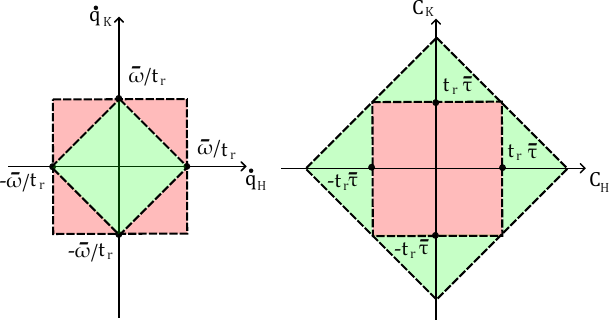}
    \caption{Description of the allowed speeds and torques at the joint level for the presented differential transmission. The red squares represent torques and speed allowed by traditional 1-to-1 actuation systems. The green ones represent those allowed by our differential solution. $\overline \omega$ and $\overline{\tau}$ represent the maximum motor speed and torque, respectively.}
    \label{fig:transmission_domain}
\end{figure}

This coupled transmission is, in general, able to produce a broader range of torques on the joints with respect to traditional solutions with independent actuation. Indeed, if the user is executing a task particularly demanding for a joint but not for the other, the exoskeleton can direct the power from both motors on that joint. This comes along with a disadvantage on the allowed speeds because, with respect to a classical one-to-one transmission system, our solutions permit lower joint speeds. This is visually explained in  Fig.~\ref{fig:transmission_domain}.

Thanks to this coupling, both peak (\SI{453}{Nm}) and nominal (\SI{192}{Nm}) torques at the hip and knee joints can be achieved by summing the contributions of the two motors. However, the coupling requires checking that both required torques at the knee and hip joints are achievable, i.e., the required $(C_H,C_K)$  lies in the colored area for the plot shown in Fig.~\ref{fig:transmission_domain}.

\section{Friction Model}
\label{sec:III}

Supposing, for the sake of simplicity,  a unitary transmission ratio $t_r=1$, the mechanical transmission presented in this study is kinematically and statically equivalent to the scheme depicted in Fig.~\ref{fig:transmission_scheme_model}. 
There, the top and bottom horizontal bars represent the transmission systems of the two motors, while the vertical oval bars correspond to the hip and knee joints, attached to the respective thigh and shin links, which are represented respectively by the central points of the two oval bars. Moreover, the two T-shaped bars represent the links that act as references to the various transmission segments.

We propose a formulation that consists of applying a physically-grounded friction model at the level of the individual branches of a coupled differential transmission, rather than at the equivalent joint level. This allows the different operating conditions of each segment to be explicitly considered.
Following the positive directions given by the arrows, we can see that, at the joint level, the speed output  is 
\begin{equation}
\left\{    
\begin{aligned}
    \dot q_H &= \frac{\dot \theta_1 + \dot \theta_2}{2}\\
    \dot q_K &= \frac{\dot \theta_1 - \dot \theta_2}{2}
\end{aligned}
\right.
\end{equation}
which is the same as in Eq.~\ref{eq:transm_matrix_vel}, while the torque output is
\begin{equation}\label{eq:jnt_tau_friction}
\left\{
\begin{aligned}
    C_H &= \tau_{1H} + \tau_{2H}\\
    C_K &= \tau_{1K} - \tau_{2K}
\end{aligned}
\right.
\end{equation}
which is almost the same as in Eq.~\ref{eq:transm_matrix_vel}, with the difference that here $\tau_{1H}, \tau_{2H}, \tau_{1K},$ and $\tau_{2K}$ take into account losses due to friction.

To quantitatively evaluate the friction compensation torques of an actuation system, it is necessary to define an appropriate friction model that can be used under all possible operating conditions of the actuation system. The evaluation of the compensation torques can be carried out quantitatively only when the components of the system move with respect to their reference frame. 
Moving from the Coulomb model, in which friction depends on the sign of the speed, with an eventual viscous term (like in \cite{devzman2022mechanical}), we approach compensation during motion and propose a model which assumes the existence of two distinct contributions.
The first contribution is independent of the net load acting on the transmission.
It accounts for dry friction phenomena and for the preload in the cables. The second contribution depends explicitly on the transmitted torque, similarly to the approach adopted in \cite{hassen2021exoskeleton} and captures load-dependent losses in the supports, which are implemented using rolling bearings. For this term, a static linear dependence was chosen instead of more complex dynamic models, such as the LuGre model \cite{olsson1998friction}, which are difficult to identify reliably, especially in the presence of a differential transmission. 

With reference to Fig.~\ref{fig:transmission_scheme_model}, the proposed friction model for each transmission stage is the following:
\begin{equation}\label{eq:friction_model}
\left\{
\begin{aligned}
    &\tau_{out} = \tau_{in}  - F\ \text{sign}\ (\omega)\\ 
    &F = 2\mu \frac{f_0}{1+\mu} + 2\frac{1-\mu}{1+\mu}\frac{||\tau_{in}+\tau_{out}||}{2}=\\
    &\quad= \bar{f}_0 + \bar{\mu} \frac{||\tau_{in}+\tau_{out}||}{2} 
\end{aligned}
\right.
\end{equation}

where $\tau_{in}$ and $\tau_{out}$ are the torques before and after the specific segment, $\omega$ is its speed, and $F$ is the friction torque, which is composed of a load-independent term, i.e. $\bar{f}_0$, and a load-dependent term. The load-dependent contribution is, for symmetry reasons, dependent on both the input and output torques. 
The average magnitude of the input and output torques is adopted as a symmetric approximation of the effective transmitted load, avoiding privileging either side of the stage.
In Fig.~\ref{fig:transmission_scheme_model}, the reference speed for the stages 1B and 2B is set to the hip joint speed, since the bearings are supported by the thigh link. In contrast, for stages 1A and 2A, the bearings are supported by the same link to which the motor stators are fixed, so the reference speed is zero.

This model can then be declined in four possible cases, depending on the signs of the various components.
When $\tau_{in}+\tau_{out}>0$ and $\omega>0$, Eq.~\ref{eq:friction_model} becomes
\begin{equation}\label{eq:direct_motion}
    \tau_{out} = \mu (\tau_{in}-f_0)
\end{equation}
and the same happens in case $\tau_{in}+ \tau_{out}<0$ and $\omega<0$. These correspond to the case of \textit{direct motion} of a transmission segment. 
When $\tau_{in}+\tau_{out}>0$ and $\omega<0$, or $\tau_{in}+\tau_{out}<0$ and $\dot \theta>0$, Eq.~\ref{eq:friction_model} becomes
\begin{equation}\label{eq:reverse_motion}
    \tau_{out} = \tau_{in}/\mu-f_0
\end{equation}
This is the equation corresponding to the case of \textit{reverse motion} of a specific transmission segment.

\section{Identification procedure}
\label{sec:IV}

\begin{figure}
    \centering
    \includegraphics[width=0.75\columnwidth]{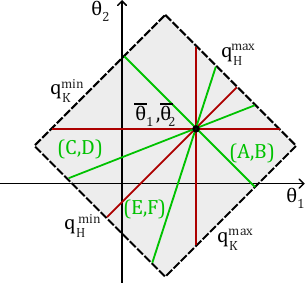}
    \caption{The workspace in the motor space allowed by the device mechanics. The point $(\overline{\theta}_1,\overline{\theta}_2)$ represents the starting point of the exploratory movements. The dotted segments represent the points where the device's mechanical stops are reached for the HFE ($q_3^{max}$ and $q_3^{min}$) or the KFE ($q_4^{max}$ and $q_4^{min}$). Given the starting point $(\overline \theta_1, \overline \theta_2)$, the red segments represent the movements for which there is no movement of some part of the whole mechanical transmission ($\dot \theta_1 =0$, $\dot \theta_2 = 0$, or $\dot \theta_1 = \dot \theta_2$), while the green ones are the exploratory movements we chose for the friction parameters identification.}
    \label{fig:transmission_workspace}
\end{figure}

For the identification of the proposed model, we need to guarantee that all the transmission segments move during data collection.
Given the schematic model for the friction of the transmission, and the complete workspace of the possible motor positions allowed by the mechanics (see Fig.~\ref{fig:transmission_workspace}), we decided to realize six motor movements which permit us to write six equations in the eight unknown friction parameters, i.e., $f_0$ and $\mu$ for each transmission segment.

\begin{table}
   \centering
   \caption{Operative conditions for each exploratory trajectory.}
   \label{tab:traj_motion_condition}
   \resizebox{0.9\columnwidth}{!}{%
   \begin{tabular}{c | c | c | c | c | c }
   \hline\hline
        Traj & Condition & 1A & 1B & 2A  & 2B \\
   \hline
        A & $\dot \theta_1 = -\dot \theta_2 <0$ & direct & direct & reverse & reverse \\
        B & $\dot \theta_1 = -\dot \theta_2>0$ & reverse & reverse & direct & direct \\           
        C & $\dot \theta_1 = 2\dot \theta_2 >0$ & direct & direct & direct & reverse \\
        D & $\dot \theta_1 = 2\dot \theta_2<0$ & reverse & reverse & reverse & direct \\           
        E & $\dot \theta_2 = -2\dot \theta_1>0$ & direct & reverse & direct & direct \\
        F & $\dot \theta_2 = -2\dot \theta_1<0$ & reverse & direct & reverse & reverse \\   
    \hline \hline
   \end{tabular}%
   }
\end{table}

We chose these three pairs of trajectories to be as excitatory as possible for the system. In fact, \textit{assuming motor torques to be positive}, they represent all the achievable combinations of direct and reverse motion for the transmission segments (see Tab.~\ref{tab:traj_motion_condition}). This is because the red segments divide the workspace into independent operative zones, so we took one pair of trajectories for each pair of diametrically opposite zones. 

The \textbf{first direction} we explore is the $(A,B)$ one. 

We thus realize a manoeuvre such that ${\dot \theta_1 = -\dot \theta_2 > 0}$ (manouevre A). 
For the first transmission, since both segments are in direct motion, we can apply Eq.~\ref{eq:direct_motion}, obtaining:
\begin{align*}
    \tau_{1H}^A &= (\tau_1^A-f_{1A})\mu_{1A},\\
    \tau_{1K}^A &= (\tau_{1H}^A-f_{1B})\mu_{1B} =
    ((\tau_1^A-f_{1A})\mu_{1A}-f_{1B})\mu_{1B}.
\end{align*}
For the second transmission, since both segments are in reverse motion, we can apply Eq.~\ref{eq:reverse_motion}, obtaining:
\begin{align*}
    &\tau_{2H}^A = \frac{\tau_{2}^A}{\mu_{2A}} + f_{2A}\\
    &\tau_{2K}^A = \frac{\tau_{2H}^A}{\mu_{2B}} + f_{2B} = \left(\frac{\tau_{2}^A}{\mu_{2A}} + f_{2A}\right)\frac{1}{\mu_{2B}} + f_{2B}\\
\end{align*}
Therefore, on each joint we have
\begin{align*}
    C_H^A &= \tau_{1H}^A + \tau_{2H}^A = 
    (\tau_{1}^A-f_{1A})\mu_{1A} + 
    \frac{\tau_{2}^A}{\mu_{2A}} + f_{2A}\\
    C_K^A &= \tau_{1K}^A - \tau_{2K}^A = ((\tau_1^A-f_{1A})\mu_{1A}-f_{1B})\mu_{1B} + ...\\ 
    & \quad \ - \left(\frac{\tau_{2}^A}{\mu_{2A}} + f_{2A}\right)\frac{1}{\mu_{2B}} - f_{2B}.
\end{align*}

We then realize a manoeuvre such that $\dot \theta_1 = -\dot \theta_2 < 0$ (manouevre B). 
For the first transmission, since both segments are in reverse motion, we can apply Eq.~\ref{eq:reverse_motion}, obtaining:
\begin{align*}
    &\tau_{1H}^B = \frac{\tau_{1}^B}{\mu_{1A}}+f_{1A},\\
    &\tau_{1K}^B = \frac{\tau_{1H}^B}{\mu_{1B}}+f_{1B} = \left(\frac{\tau_{1}^B}{\mu_{1A}}+f_{1A}\right)\frac{1}{\mu_{1B}}+f_{1B} .
\end{align*}
For the second transmission, sinceboth segments are in direct motion, we can apply Eq.~\ref{eq:direct_motion}, obtaining:
\begin{align*}
    &\tau_{2H}^B = (\tau_{2}^B - f_{2A})\mu_{2A},\\
    &\tau_{2K}^B = (\tau_{2H}^B - f_{2B})\mu_{2B} =  ((\tau_{2}^B - f_{2A})\mu_{2A} - f_{2B})\mu_{2B}.
\end{align*}
Thus, on each joint, we have
\begin{align*}
    C_H^B &= \tau_{1H}^B + \tau_{2H}^B = \frac{\tau_{1}^B}{\mu_{1A}}+f_{1A} + (\tau_{2}^B - f_{2A})\mu_{2A},\\
    C_K^B &= \tau_{1K}^B - \tau_{2K}^B = \left(\frac{\tau_{1}^B}{\mu_{1A}}+f_{1A}\right)\frac{1}{\mu_{1B}}+f_{1B} + ... \\
    &\quad \  -((\tau_{2}^B - f_{2A})\mu_{2A} - f_{2B})\mu_{2B}.
\end{align*}

Finally, if we consider each point $(\theta_1, \theta_2)$ through which we passed during both explorations A and B, we will have $C_H^A(\theta_1, \theta_2) = C_H^B(\theta_1, \theta_2)$ and $C_K^A(\theta_1, \theta_2) = C_K^B(\theta_1, \theta_2)$:
\begin{multline}\label{eq:fri_eff_1H}
    ((\tau_{1}^A-f_{1A})\mu_{1A} + \frac{\tau_{2}^A}{\mu_{2A}} + f_{2A} = \\ 
    =\frac{\tau_{1}^B}{\mu_{1A}}+f_{1A} + (\tau_{2}^B - f_{2A})\mu_{2A},
\end{multline}
\begin{multline}\label{eq:fri_eff_1K}
     ((\tau_1^A-f_{1A})\mu_{1A}-f_{1B})\mu_{1B} - \left(\frac{\tau_{2}^A}{\mu_{2A}} + f_{2A}\right)\frac{1}{\mu_{2B}} - f_{2B} = \\
    =\left(\frac{\tau_{1}^B}{\mu_{1A}}+f_{1A}\right)\frac{1}{\mu_{1B}}+f_{1B} - ((\tau_{2}^B - f_{2A})\mu_{2A} - f_{2B})\mu_{2B}.
\end{multline}

The \textbf{second direction} we explored is the $(C,D)$ one.
We first realize a manoeuvre such that $\dot \theta_1 > \dot \theta_2 > 0$ (manouevre C) and then realize a manoeuvre such that $\dot \theta_1 <\dot \theta_2 < 0$ (manouevre D). 
Following steps similar to those shown for trajectories A and B, if we consider each point $(\theta_1, \theta_2)$ through which we passed during both explorations A and B, we will have $C_H^C(\theta_1, \theta_2) = C_H^D(\theta_1, \theta_2)$ and $C_K^C(\theta_1, \theta_2) = C_K^D(\theta_1, \theta_2)$:
\begin{multline}\label{eq:fri_eff_2H}
    (\tau_{1}^C-f_{1A})\mu_{1A} + (\tau_{2}^C-f_{2A})\mu_{2A} = \\ 
    =\frac{\tau_{1}^D}{\mu_{1A}}+f_{1A} + \frac{\tau_{2}^D}{\mu_{2A}} + f_{2A},
\end{multline}
\begin{multline}\label{eq:fri_eff_2K}
     ((\tau_1^C-f_{1A})\mu_{1A}-f_{1B})\mu_{1B} - ((\tau_2^C-f_{2A})\frac{\mu_{2A}}{\mu_{2B}} - f_{2B} = \\
    =\left(\frac{\tau_{1}^D}{\mu_{1A}}+f_{1A}\right)\frac{1}{\mu_{1B}}+f_{1B} - \left(\frac{\tau_{2}^D}{\mu_{2A}} + f_{2A}-f_{2B}\right)\mu_{2B}.
\end{multline}

The \textbf{third direction} we explored is the $(E,F)$ one.
We first realize a manoeuvre such that $\dot \theta_2 > \dot \theta_1 > 0$ (manouevre E) and then realize a manoeuvre such that $\dot \theta_2 < \dot \theta_1 < 0$ (manouevre F). 
Again, following steps similar to those shown for trajectories A and B, if we consider each point $(\theta_1, \theta_2)$ through which we passed during both explorations A and B, we will have $C_H^E(\theta_1, \theta_2) = C_H^F(\theta_1, \theta_2)$ and $C_K^E(\theta_1, \theta_2) = C_K^F(\theta_1, \theta_2)$:
\begin{multline}\label{eq:fri_eff_3H}
    (\tau_{1}^E-f_{1A})\mu_{1A} + (\tau_{2}^E-f_{2A})\mu_{2A} = \\ 
    =\frac{\tau_{1}^F}{\mu_{1A}}+f_{1A} + \frac{\tau_{2}^F}{\mu_{2A}} + f_{2A},
\end{multline}
\begin{multline}\label{eq:fri_eff_3K}
     (\tau_1^E-f_{1A})\frac{\mu_{1A}}{\mu_{1B}}+f_{1B} - ((\tau_2^E-f_{2A})\mu_{2A} - f_{2B})\mu_{2B} = \\
    =\left(\frac{\tau_{1}^F}{\mu_{1A}}+f_{1A}-f_{1B}\right)\mu_{1B} - \left(\frac{\tau_{2}^F}{\mu_{2A}} + f_{2A}\right)\frac{1}{\mu_{2B}}-f_{2B}.
\end{multline}

At this point, if we put together Eq.~\ref{eq:fri_eff_1H}-\ref{eq:fri_eff_3K}, we have six independent equations in eight unknowns. However, since each equation can be computed at various points, an optimization procedure can be used to find the best-fitting friction parameters.

\section{Experimental Results}
\label{sec:V}

We tested the friction parameter identification procedure described in the previous Section to verify its validity, and obtained the following values:
\[
\begin{array}{c c c c c c c c}
f_{1A} & \mu_{1A} & f_{2A} & \mu_{2A} & f_{1B} & \mu_{1B} & f_{2B} & \mu_{2B} \\
0.089   & 0.90      & 0.077   & 0.91      & 0.001   & 0.99     & 0.014   & 0.97
\end{array}
\]
where $f_{1A}, f_{2A}, f_{1B}, \text{ and }f_{2B}$ are expressed in Nm, while the others are dimensionless numbers.

\begin{figure}
    \centering
    \includegraphics[width=0.95\columnwidth]{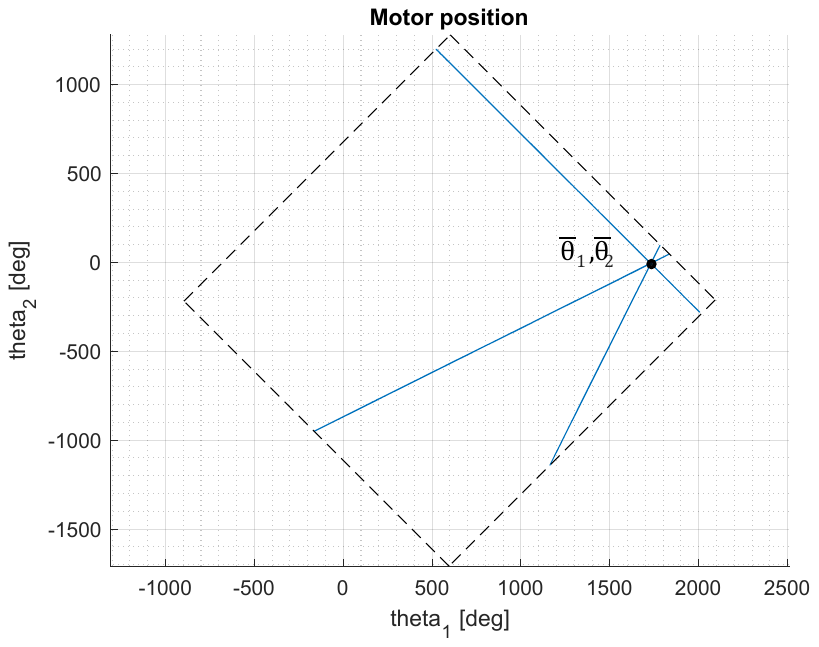}
    \caption{Actual motor trajectories during the 6 exploratory movements. The dotted segments represent the device’s mechanical stops. The starting point $(\overline{\theta}_1,\overline{\theta}_2)$ is different from the one in Fig.~\ref{fig:transmission_workspace} since it depends on the hip and knee configuration: in this particular case, they were both close to their maximum values.}
    \label{fig:motor_evolution}
\end{figure}

In Fig.~\ref{fig:motor_evolution}, the actual evolution of each motor's position during exploratory trajectories is shown.
\begin{figure}
    \centering
    \includegraphics[width=1\columnwidth]{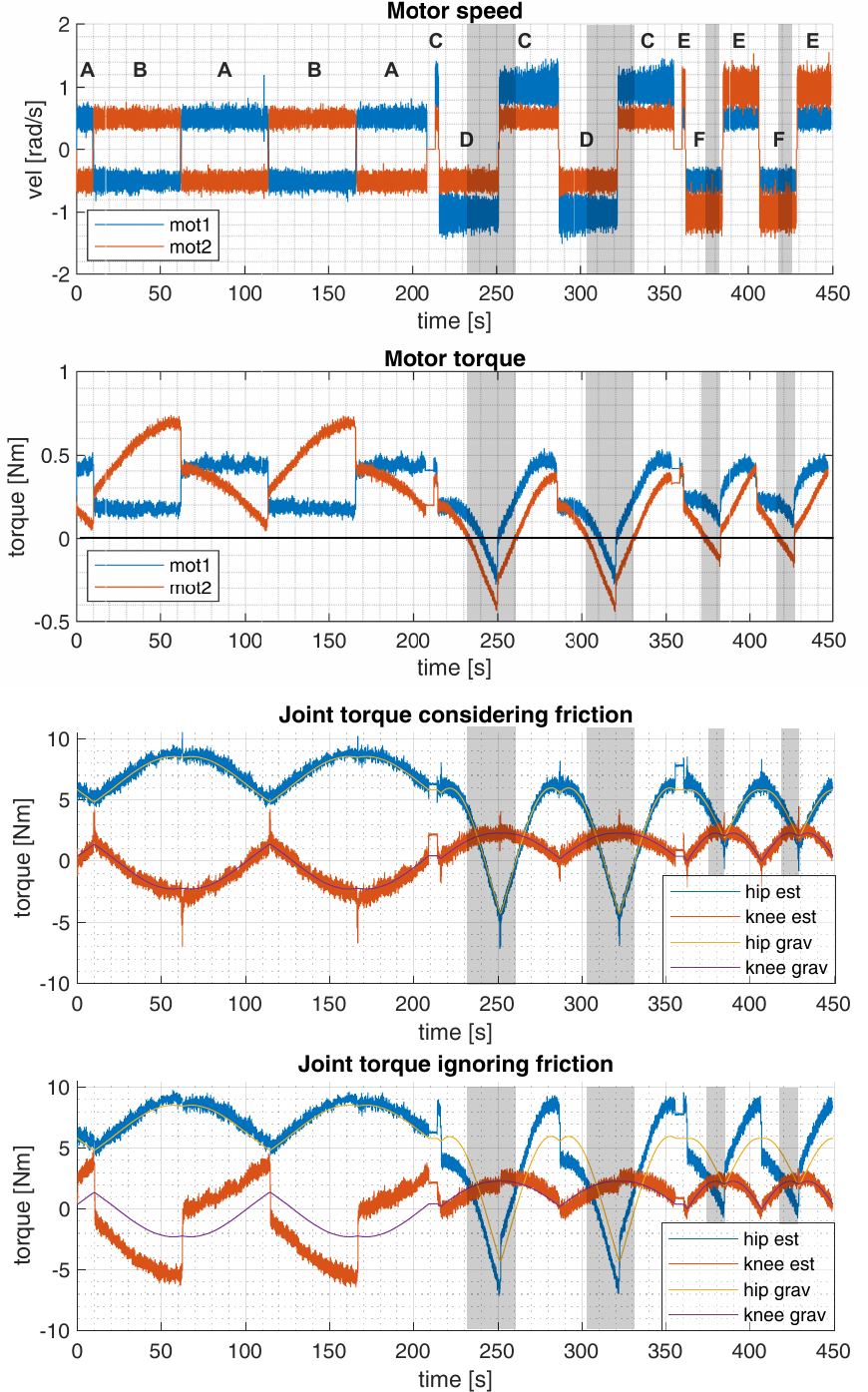}
    \caption{Result of the fitting procedure described in this study. Together with the motor speed plot, letters A-F represent which trajectory was being executed. The grey areas (corresponding to points where at least one of the two motor torques was negative) represent configurations not taken into account during estimation.}
    \label{fig:fitting_result}
\end{figure}
In Fig.~\ref{fig:fitting_result}, the quality of the fitting values is shown. We highlight how each one of the manoeuvres A-F was not executed once but at least twice, always returning to the starting point. This was done to better quantify the involved torques.
Moreover, always with reference to Fig.~\ref{fig:fitting_result},  the areas not used for the fitting, since at least one of the two motors has a negative torque, are highlighted in grey. These sections are nonetheless used to evaluate the quality of the fitting. In fact, in the bottom plot we show the estimated joint torque using Eq.~\ref{eq:transm_matrix_tau}, while in the third we can see those estimated using Eq.~\ref{eq:jnt_tau_friction}  along with the friction model Eq.~\ref{eq:friction_model}. 
Both these plots also show the actual joint torque. Since the exploratory movements are performed at constant and low speed, inertia effects can be neglected, and the only torque component is the gravity, which can be estimated from joint angles and physical parameters, such as masses and centres of gravity. 

This gravitational torque is not used in the identification process, but is instead employed as a ground truth reference to assess the accuracy of the estimated joint torque. This choice was motivated by the goal of defining an identification procedure that does not rely on prior knowledge of the gravitational model of the system, and that could therefore be applied even when such information is unavailable or unreliable.

To quantify the benefit in the reduction of the net torque estimation error, we chose to compute the RMSE and the normalized RMSE (NRMSE) between the estimated joint torque and the gravitational one, both for the hip and the knee (see Tab~\ref{tab:rmse}).

\begin{table}
   \centering
   \caption{RMSE and NRMSE between the estimated joint torque and the gravitational one, using either Eq.~\ref{eq:transm_matrix_tau} or Eq.~\ref{eq:jnt_tau_friction} along with the friction model Eq.~\ref{eq:friction_model}.}
   \label{tab:rmse}
   \begin{tabular}{c | c | c | c | c }
   \hline\hline
        \multirow{2}{*}{Eq.} & \multicolumn{2}{c|}{Hip} & \multicolumn{2}{c}{Knee}  \\
        ~ & RMSE & NRMSE & RMSE & NRMSE\\
        \hline
        \ref{eq:transm_matrix_tau} & \SI{1.54}{Nm} & 12\% & \SI{1.89}{Nm} & 41\%  \\
        \ref{eq:jnt_tau_friction}-\ref{eq:friction_model}& \SI{0.36}{Nm} & 2.8\% & \SI{0.38}{Nm} & 8.3\% \\           
        
    \hline \hline
   \end{tabular}%
\end{table}
\section{Discussion}
\label{sec:VI}

\subsection{Friction Estimation}
\label{sec:discussion_features}

The values we obtained for the various friction parameters are in line with our expectations. 
More in particular, the static frictions of branches 1A and 2A are greater than those of branches 1B and 2B, and this is because the breakaway torque of each motor, which is of the order of tens of mNm, is included in these values.
Moreover, the values of the load-dependent friction values are worse  for branches 1A and 2A, and this is because they present two gear transmissions each, which are less efficient due to the intrinsic losses caused by tooth contact (sliding friction).

Fig.~\ref{fig:fitting_result} shows how the friction model presented in this study, along with the identification procedure proposed, can reconstruct the actual net joint torque.
On the contrary, ignoring this contribution leads to noticeable errors.

\subsection{Limitations and Future Works}
\label{sec:limitation_future}
A major limitation of the proposed approach lies in the assumption of motion within the transmission. When motor speeds approach zero, or in the condition $\dot \theta_1=\dot \theta_2$, the direction of motion is not well defined for all transmission segments, and the proposed friction model cannot reliably estimate the friction losses. This causes, for example, the torque estimation errors observed during motion reversals in Fig.~\ref{fig:fitting_result}. This limitation is inherent to friction estimation strategies that rely on velocity-dependent formulations and motivates the need for specific treatments of static conditions.

As future work, the identified friction model will be simplified for the computation of the required input torque $\tau_{in}$ given a desired output torque $\tau_{out}$. In particular, Eq.~\ref{eq:friction_model} will be reformulated by considering only the output torque within the load-dependent friction term, thus avoiding the need to solve an equation containing the unknown variable (i.e., the input torque $\tau_{in}$) inside the absolute value. An analysis of the estimation accuracy using this simplification will be presented. 

Moreover, the approach used for the real-time implementation of a friction \textit{compensation} strategy based on the friction model presented in this study will also be presented.

Within a complete exoskeleton framework, the proposed module can be integrated as the low-level actuation layer, while higher-level controllers determine the desired assistance profile during gait. In this context, future studies will investigate how torque is effectively distributed between hip and knee joints under different tasks and load conditions.

Finally, the proposed sensorless torque estimation strategy will be validated during experiments with human users. To quantitatively assess transparency and torque estimation performance, joint-level torque sensors will be temporarily integrated and used as ground truth, enabling a direct comparison between estimated and measured joint torques.

\section{Conclusion}
\label{sec:VII}

This paper presented a novel and innovative actuation system for lower-limb exoskeletons consisting of a differential transmission featuring a coupling between the hip and knee flexion-extensions. The system design and friction characterization strategy have been described in detail, leading to an experimental assessment of the proposed identification method.

\bibliographystyle{elsarticle-num}
\bibliography{./Bibliography/MyBiblio}

\begin{thebibliography}{10}
\expandafter\ifx\csname url\endcsname\relax
  \def\url#1{\texttt{#1}}\fi
\expandafter\ifx\csname urlprefix\endcsname\relax\def\urlprefix{URL }\fi
\expandafter\ifx\csname href\endcsname\relax
  \def\href#1#2{#2} \def\path#1{#1}\fi

\bibitem{zhou2021lower}
J.~Zhou, S.~Yang, Q.~Xue, Lower limb rehabilitation exoskeleton robot: A
  review, Advances in Mechanical Engineering 13~(4) (2021) 16878140211011862.

\bibitem{kuber2023systematic}
P.~M. Kuber, M.~M. Alemi, E.~Rashedi, A systematic review on lower-limb
  industrial exoskeletons: Evaluation methods, evidence, and future directions,
  Annals of biomedical engineering 51~(8) (2023) 1665--1682.

\bibitem{jia2020preliminary}
Z.~Jia-Yong, L.~Ye, M.~Xin-Min, H.~Chong-Wei, M.~Xiao-Jing, L.~Qiang,
  W.~Yue-Jin, Z.~Ang, A preliminary study of the military applications and
  future of individual exoskeletons, in: Journal of Physics: Conference Series,
  Vol. 1507, IOP Publishing, 2020, p. 102044.

\bibitem{bettella2025scoping}
F.~Bettella, S.~Tortora, E.~Menegatti, N.~Petrone, A.~Del~Felice, A scoping
  review on lower limb exoskeleton actuation’s description and
  characteristics, Robotica (2025) 1--18.

\bibitem{choi2019control}
B.~Choi, C.~Seo, S.~Lee, B.~Kim, Control of power-augmenting lower extremity
  exoskeleton while walking with heavy payload, International Journal of
  Advanced Robotic Systems 16~(1) (2019) 1729881419830535.

\bibitem{yu2023design}
L.~Yu, H.~Leto, S.~Bai, Design and gait control of an active lower limb
  exoskeleton for walking assistance, Machines 11~(9) (2023) 864.

\bibitem{liu2024synergetic}
H.~Liu, C.~Zhu, Z.~Zhou, Y.~Dong, W.~Meng, Q.~Liu, Synergetic gait prediction
  and compliant control of sea-driven knee exoskeleton for gait rehabilitation,
  Frontiers in Bioengineering and Biotechnology 12 (2024) 1358022.

\bibitem{yu2020quasi}
S.~Yu, T.-H. Huang, X.~Yang, C.~Jiao, J.~Yang, Y.~Chen, J.~Yi, H.~Su,
  Quasi-direct drive actuation for a lightweight hip exoskeleton with high
  backdrivability and high bandwidth, IEEE/ASME Transactions On Mechatronics
  25~(4) (2020) 1794--1802.

\bibitem{witte2020design}
K.~A. Witte, S.~H. Collins, Design of lower-limb exoskeletons and emulator
  systems, in: Wearable robotics, Elsevier, 2020, pp. 251--274.

\bibitem{sun2025design}
Q.~Sun, W.~Feng, H.~Chen, Z.~Wang, L.~Cui, W.~Ren, Design and performance
  evaluation of a knee assist exoskeleton based on bowden cables transmission,
  in: 2025 9th International Conference on Robotics, Control and Automation
  (ICRCA), IEEE, 2025, pp. 11--17.

\bibitem{xu2025survey}
J.~Xu, S.~Chen, S.~Li, Y.~Liu, H.~Wan, Z.~Xu, C.~Zhang, A survey on design and
  control methodologies of high-torque-density joints for compliant lower-limb
  exoskeleton, Sensors 25~(13) (2025) 4016.

\bibitem{shi2022learning}
Y.~Shi, M.~Guo, C.~Hui, S.~Li, X.~Ji, Y.~Yang, X.~Luo, D.~Xia, Learning-based
  repetitive control of a bowden-cable-actuated exoskeleton with frictional
  hysteresis, Micromachines 13~(10) (2022) 1674.

\bibitem{chishty2021kinematic}
H.~A. Chishty, A.~Zonnino, A.~J. Farrens, F.~Sergi, Kinematic compatibility of
  a wrist robot with cable differential actuation: Effects of misalignment
  compensation via passive joints, IEEE transactions on medical robotics and
  bionics 3~(4) (2021) 970--979.

\bibitem{chen2019elbow}
T.~Chen, R.~Casas, P.~S. Lum, An elbow exoskeleton for upper limb
  rehabilitation with series elastic actuator and cable-driven differential,
  IEEE Transactions on Robotics 35~(6) (2019) 1464--1474.

\bibitem{devzman2022mechanical}
M.~De{\v{z}}man, T.~Asfour, A.~Ude, A.~Gams, Mechanical design and friction
  modelling of a cable-driven upper-limb exoskeleton, Mechanism and Machine
  Theory 171 (2022) 104746.

\bibitem{andrade2021role}
R.~M. Andrade, P.~Bonato, The role played by mass, friction, and inertia on the
  driving torques of lower-limb gait training exoskeletons, IEEE Transactions
  on Medical Robotics and Bionics 3~(1) (2021) 125--136.

\bibitem{hassen2021exoskeleton}
M.~D. Hassen, I.~Laamiri, H.~Messaoud, Exoskeleton dynamic modeling and
  identification with load and temperature-dependent friction model, in: 2021
  IEEE 2nd International Conference on Signal, Control and Communication (SCC),
  IEEE, 2021, pp. 14--19.

\bibitem{baud2021review}
R.~Baud, A.~R. Manzoori, A.~Ijspeert, M.~Bouri, Review of control strategies
  for lower-limb exoskeletons to assist gait, Journal of neuroengineering and
  rehabilitation 18~(1) (2021) 119.

\bibitem{ALEx}
{ALEx RS by Wearable Robotics s.r.l.},
  \url{https://nexumrobotics.it/solutions/alex-rs/}, {Last accessed on
  2025-09-26}.

\bibitem{olsson1998friction}
H.~Olsson, K.~J. {\AA}str{\"o}m, C.~C. De~Wit, M.~G{\"a}fvert, P.~Lischinsky,
  Friction models and friction compensation, Eur. J. Control 4~(3) (1998)
  176--195.

\end{thebibliography}

\end{document}